# Explainable Mixed Data Representation and Lossless Visualization Toolkit for Knowledge Discovery

Boris Kovalerchuk
*Dept. of Computer Science*
*Central Washington University*
USA
BorisK@cwu.edu

Elijah McCoy
*Dept. of Computer Science*
*Central Washington University*
USA
Elijah.McCoy@cwu.edu

*Abstract*—Developing Machine Learning (ML) algorithms for heterogeneous/mixed data is a longstanding problem. Many ML algorithms are not applicable to mixed data, which include numeric and non-numeric data, text, graphs and so on to generate interpretable models. Another longstanding problem is developing algorithms for lossless visualization of multidimensional mixed data. The further progress in ML heavily depends on success interpretable ML algorithms for mixed data and lossless interpretable visualization of multidimensional data. The later allows developing interpretable ML models using visual knowledge discovery by end-users, who can bring valuable domain knowledge which is absent in the training data. The challenges for mixed data include: (1) generating numeric coding schemes for non-numeric attributes for numeric ML algorithms to provide accurate and interpretable ML models, (2) generating methods for lossless visualization of n-D non-numeric data and visual rule discovery in these visualizations. This paper presents a classification of mixed data types, analyzes their importance for ML and present the developed experimental toolkit to deal with mixed data. It combines the Data Types Editor, VisCanvas data visualization and rule discovery system which is available on GitHub.

## I. Introduction

Developing Machine Learning (ML) algorithms for **heterogeneous/mixed data** is a longstanding problem. Very many current ML algorithms including Deep Learning algorithms are not designed for discovering models on heterogeneous data. Heterogeneous data include data of multiple types, including numeric and non-numeric data, text, graphs and so on. Often non-numeric data are *nominal* data with values like red, green, and blue, or *ordinal* (*ordered*) data like large, medium, and small. Another longstanding problem is developing algorithms for **visualization** of multidimensional heterogeneous data [11] and for **lossless** visualization specifically [10]. The literature in this domain focuses on visualization of nominal, ordinal, and numeric attributes in parallel coordinates [1] but without link to interpretable ML.

The further progress in ML heavily depends on both success interpretable ML algorithms for heterogenous data and lossless interpretable visualization of multidimensional data. The later allows developing interpretable ML models using visual knowledge discovery [13] by end-users, who can bring valuable domain knowledge which is absent in the training data. Moreover, often interpretability and appeal of *visual ML models* are higher for the end user than interpretability of the analytical ML models. This opens a new opportunity for expanding and strengthening the domain of interpretable ML.

A common approach to make existing numeric ML algorithms applicable to non-numeric data is numeric coding these data [4-6]. If coding corrupts properties of these attributes, like value similarities, it leads to **non-interpretable ML models** for kNN and other algorithms based on the distances, which already happened with a popular integer coding. Unfortunately, the attention to interpretability in the coding literature is limited while it is as a major roadblock to widen ML use with mixed data in the domains with the high cost of errors like medicine.

Thus, the challenges are:
(1) Developing numeric **coding schemes** for non-numeric attributes for ML algorithms to provide **accurate** and **interpretable** ML models for heterogeneous data.
(2) Developing methods for lossless **visualization** of n-D **non-numeric** data and **visual rule discovery** in these visualizations.

This paper is organized as follows. Section II proposes a classification of mixed data types and analyzes their import role in Machine Learning. Section III presents measurement data types and introduces the Toolkit. Section VI describes the proposed approach. Section V describes data visualization in the Toolkit, and Section VI summarizes the results and outlines the future work.

## II. Heterogeneous data types and their role in Machine Learning

This section provides a classification of mixed data types with comparative analyses of their importance for Machine Learning. In includes a description of several coding schemas of measurement data types.

### A. Classification of mixed data types for machine learning

Heterogeneity/mixture of data is not limited by numeric, nominal, and ordinal data types that are **measurement data types** [16]. Numeric data are not homogeneous, but heterogeneous too. They have several measurement types such as **absolute**, **ratio, interval**, and **cyclical**. The strongest one (**absolute**) has the largest set of meaningful arithmetic relations and operations, which ML algorithms can conduct with them to produce interpretable models. This data type includes attributes that represent the counting the number of elements which has unique not arbitrary 0 and unit 1. The next type is **ratio** data type, which has a unique zero, but not a unit, like Kelvin temperature with unique physical zero, but the unit is defined by convention. Other data types include **interval** data type (e.g., Celsius and Fahrenheit temperature where both zero and the unit are defined by convention) and values can be negative, **cyclical** (e.g., azimuth) [15] and others. These data types have meaningful arithmetic operations, distances, frequency distributions, different averages, standard deviation, and others, which ML algorithms can use to get interpretable models.

However, the meaningful difference between two values in the cyclical azimuth attribute between $1°$ and $359°$ is $2°$ not 358 as it is for the attribute like weight. In contrast the **ordinal data** allow only checking only relations like $a \leq b$ without allowing arithmetic operations. Thus, ML algorithms, which do these operations with ordinal data will produce non-interpretable models. Finally **nominal** data will allow ML algorithms only testing $a \leq b$. All other operations will lead to non-interpretable ML models. Graphs and texts are other examples of data heterogeneity, which are not given by a set of well-defined attributes. The attributes need to be made from them. We will call such data as **non-attribute-based data types**.

Another category of data heterogeneity is **real-world physical modality**. Both weight and height are ratio attributes. They have physical zeros of weight and height, but arbitrary units. It is important that they have *different real-world physical meaning, modalities*. Their sum and other arithmetic operations have *no meaning*, e.g., $3kg + 5m$. We will call such data types as **modality data** types. This makes popular linear machine learning models non-interpretable for data of different modalities [18], while they can be interpretable for data of the same modality type as temperature time series.

Thus, each dataset is characterized by the several categories of **data types**: 1) type of *measurement*, (2) *attribute, non-attribute*, and (3) type of *modality*. If a dataset contains data of different such types, then this dataset is **heterogenous/mixed**. Data type differences dramatically impact accuracy and interpretability of ML models. The use of the ML algorithm for the data of types for which it was not designed leads to inaccurate and non-interpretable models.

While impact of measurement data type and attribute, non-attribute data types [4-6] are well recognized, the impact of physical modality is less recognized in ML literature, which led to several unsubstantiated claims in the literature such as claiming that linear models are always interpretable, which critically analyzed in [18]. Extensive current studies focus on producing numeric data (strings of numbers, vectors) from non-numeric data, such as graphs and texts [17] known as **embedding**. While this is a popular approach, which attempts to preserve the structure of the original data, it has a significant deficiency. Interpretability of those vectors is not clear or absent, while those models are often quite accurate.

Building interpretable models from mixed n-D data by **encoding** individual attributes according to their measurements data types [4-6] can be easier. Often the data preprocessing step coverts/encodes the original attributes to other attributes by *many-to-one mappings* to decrease the attribute space and to make attributes more relevant to the ML task. It results in **loss** of some information and can change and corrupt the data type.

For instance, the absolute numeric data in [0,100] interval can be converted to ordinal data, where code 1 is for interval [0,50), code 2 for [50,70) and code 3 for [70,100]. It means that values a=10, b=60, c=100 will be converted to codes cd(a)=1, cd(b)= 2 and cd(c)=3. In the original absolute data, c=10a and b=6a, but in codes cd(b)=2cd(a) and cd(c)=3cd(a), if we treat codes as numeric data of the same measurement type as original data. This example shows that, we cannot automatically expand the data type to a modified attribute. A domain expert should confirm that code ratios cd(b)=2cd(a) and cd(c)=3cd(a) have meaning in the domain. Otherwise, the ML model that uses them is not interpretable in that domain. Without such confirmation from a domain expert, we cannot treat codes as a numeric data type and need to treat them as ordinal data where computing ratios is not allowed and has no meaning. Next, without similar confirmation differences between cd(b)-cd(a)=2-1=1 and cd(c) -cd(a)=3-1=2 have no meaning too. As a result, ML algorithms based on distances like kNN will produce non-interpretable models here.

One of the approaches to enrich nominal attributes is computing dissimilarity between each two values of a nominal attribute relative to the *target attribute* of the ML classification task. A way to do this is minimizing an error function on the training samples taking to account correlation among attributes and producing the adaptive dissimilarity matrices [7]. This method called ADM generalizes the value difference metric (VDM) [8]. A similar idea is presented in [5] for SVM. Such approach improved the ML model accuracy and interpretations of the relationships among the different values of nominal attributes [7] for some datasets from UCI ML repository.

However, for the mushroom data [9], which we work in this paper, the lowest error of 0.4% reported in [7] was obtained without ADM and VDM, but by the interpretable C4.5 Decision Tree (DT) algorithm. This lowest error is still not 0, because DTs can miss some better alternatives considering each attribute only one at a time during node splitting [7]. The important difficulty of ADM is that it is based on a ML algorithm used to compute the error function on the training data. Another ML algorithm will lead to a different result and if the ML algorithm is non-interpretable then the ADM results also will be non-interpretable. The radial basis function (RDF) classifier was used in [7] as the ML algorithm. It sums up the exp functions of the distances. The interpretability of it for the mushroom task is not clear, in contrast with decision trees or logical rules. The result with RDF needs to be independently confirmed by the domain knowledge which was done in [7] for the odor mushroom attribute. The coding that is interpretable in the domain and established without using the target attribute does not require such domain confirmation afterward.

*B. Coding of measurement data types*

Below we briefly overview popular coding methods for **nominal, ordinal** and other attributes used in machine learning [4-6] with their relation to interpretability. In section 5, we will present expansion of these methods to fit knowledge discovery in multidimensional data using visualization.

In **One Hot Encoding,** a nominal attribute is mapped to a binary vector, where 1 indicates the presence of the given value of the attribute, and 0 indicates its absence. It significantly increases the space dimensionality and run time of ML algorithms on these extended data. This coding is meaningful because it preserves the equal Hamming distances between any two values of the nominal attribute.

The label encoding simply assigns numbers from one to ki to values of the nominal attribute. It does not preserve the equal distance for values of the nominal attribute. Thus, it is not interpretable especially for the ML algorithm that exploit the differences of distances for classification like kNN. For other methods like decision trees (DTs) which do not directly exploit distances such coding can be applied but it can produce a less efficient DTs than with other encodings.

Hashing maps categorical attributes to vectors in n-D space, where the distance between two vectors is roughly preserved. The resulting dimension is much smaller than with One Hot Encoding [4]. This is a version of a more general concept of **embedding** for multiple types of heterogeneous data. Not every machine learning method can benefit from this coding. Preserving the structure for nominal attributes means all equal distances between values. If the ML algorithms exploits the differences of the distances, then it cannot benefit from such hashing. Establishing distances for ordinal data require additional information which may not be available. It limits the use of this hashing method for these data. The next most important issue is that a new space can be not interpretable.

**Ordinal encoding** is applicable to the **ordinal data** like very short, short, medium, tall, very tall. It differs from the label coding by the fact that now numbering cannot be arbitrary, but it must follow the order of the values, e.g., 1 for very short, 2 for short, 3, for medium, 4 for tall, and 5 for very tall. The interpretable operations with ordinal data are limited by less or equal relation and should not include arithmetic operations. This limits application of many existing machine learning algorithms that use arithmetic operations.

Below we summarize statistics-based coding methods [4].
**Frequency Encoding** assigns codes to values of the attribute by their frequency. It allows exploring the link between attribute and the values of the target variable too.
**Mean Encoding** or Target Encoding computes the mean of the number of times the value of the attribute appears in the target class in the two-class classification task

**Probability Ratio Encoding** for each value uses ratio P(1)/P(0) of the frequency of this value of the attribute for class 1 to the frequency for class 2 as codes.
**James-Stein estimator** assigns as a code a weighted average of the mean target value for the observed feature value and the mean target value (regardless of the feature value).

These methods are applicable to many data types but can assign the same code to different values of the attribute making this coding lossy. We resolve this issue for visualization in Section 5.

III. PROPOSED APPROACH AND TOOLKIT

This section describes the proposed approach and Toolkit with enforcing interpretability of all internal operations of ML algorithms on heterogeneous data.

Proposed Approach

**Enforcing interpretability of all internal operations of ML algorithms**. The main concept of the proposed approach is the requirement that *all internal operations of the ML algorithm should be interpretable for producing interpretable models*. For models that deal with heterogeneous data it is a very strong limitation. For instance, for *nominal data* the algorithm cannot conduct any operation beyond checking/testing if two values are *equal or not* because all other operations are not interpretable for nominal data. This leads to algorithms, which discover **logical rules** like (1)-(3) or similar:

$$\text{If } [(x_1=a \ \& \ x_2=b) \lor (x_3=c)] \ \& \ x_4 \neq d \ \& (\neg \ x_5 \neq e)$$
$$\text{Then } \mathbf{x} \in \text{class } C_1 \quad (1)$$
$$\text{If } [(x_1=a \ \& \ x_2=b) \lor (x_3=c)] \ \& \ x_4 \neq d \ \& \ (\neg x_5 \neq e)$$
$$\text{Then } \mathbf{x} \in \text{class } C_1 \text{else } \mathbf{x} \in \text{class } C_2 \quad (2)$$
$$\text{If } [(x_1=y_1 \ \& \ x_2=y_2) \lor (x_3=c)] \ \& \ x_4 \neq y_4 \ \& \ (\neg x_5 \neq e) \& \ \mathbf{x} \in \text{class } C_1$$
$$\text{Then } \mathbf{y} \in \text{class } C_1 \quad (3)$$

for nominal n-D points $\mathbf{x}=(x_1,x_2,\ldots x_n)$, $\mathbf{y}=(y_1,y_2,\ldots y_n)$. Thus, the rules can include only logical operations $\&, \lor, \neg$ and tests if values are equal or not for one or more n-D points.

Even decision trees which are traditionally considered as interpretable ML algorithms are not formally interpretable for the nominal data because they check $\leq$ relation, which is prohibited for nominal data. Coding of nominal values by integers allows technically to apply decision trees to nominal data. However, randomness of this coding leads to different decision trees and does not guarantee finding the most accurate model. Next, the produced DT must be converted to logical rules with elimination of all $\leq$ relations and their thresholds to make it fully interpretable.

The advantage of ML algorithms, which build **logical rules** is that they are **data type universal**, i.e., can produce models that include *heterogenous* data of **all types** that are expressed in the propositional or the First order Logic (FoL) [14]. Therefore, this paper focuses on logical rules. Above rules (1) and (2) are propositional and rule 3 is an example of FoL rules.

**Addressing all heterogeneous data types**. The actual number of data types in ML tasks with mixed data is greater than what is usually listed. Consider a nominal data type for which we cannot say that **a** is closer to **b** than to **c**, for instance occupations with categories: doctors, engineers, and teachers. Next, we can add a nurse to this list and can continue to call it the nominal data type. Alternatively, we can say that doctors and nurses are closer to each other than to engineers and teachers. Then, occupations are *not nominal anymore* because more relations have meaning. Similarly, we add can technicians and teaching assistants. This creates more relations that engineers are closer to technicians and teachers are closer to teaching assistants than to other occupations. To treat these occupations as a nominal data type we can create groups: (1) doctors and nurses, (2) engineers and technicians, and (3) teachers and teaching assistants and assign codes for these groups.

While grouping allows us to go back to the nominal data, we lose important similarity information about occupations. The resulting ML model on groups can be less accurate than without grouping. Instead of grouping we can assign the following codes: nurse (1), doctor (2), technician (5), engineer (6), teaching assistant (10), teacher (11) with limiting the set of operations and relations that are considered as interpretable and allowable. For instance, relation,

$$c(doctor)-c(nurse) < c(teacher)-c(engineer) \quad (4)$$

is allowed. Here, c(.) is an integer code for the occupation, e.g., c(nurse)=1. Thus, the ML algorithm can use (4), but not (5):

$$c(doctor)-c(engineer) < c(teacher)-c(engineer) \quad (5)$$

This example shows that to get an interpretable ML model with non-numeric data like occupations we can use a numeric coding of them, but we need **to limit operations and relations** which can be conducted with these numbers by ML algorithms to ensure that algorithms will produce interpretable models. In summary, this example shows that the design of ML algorithms for mixed data needs to address many data types. The relational ML algorithms [14] fit well this task for mixed data.

**Building interpretable models based on the interpretable atoms (hyperblocks).** The next important concept in our approach for interpretable ML on heterogenous data is the concept of the n-D data hyperblocks (HBs). Some HBs serve as *interpretable data atoms*.

Definition. A **numeric hyper-block** (hyperrectangle, n-orthotope) is a set of numeric n-D points $\{\mathbf{x}=(x_1,x_2,\ldots,x_n)\}$ with **center** in n-D point $\mathbf{c}=(c_1,c_2,\ldots,c_n)$ and *lengths* $\mathbf{L}=(L_1, L_2,\ldots, L_n)$ such that

$$\forall i \in I_u \; \|x_i-c_i\| \leq L_i/2 \quad (6)$$

This definition assumes that attributes a numeric and the difference between values is meaningful. This means that data are of the interval data type at least. For heterogenous data we need another definition which will include different data types. We first define the hyper-block for ordinal data and then for nominal data.

Definition. An **ordinal hyper-block** (hyperrectangle, n-orthotope) is a set of ordinal n-D points $\{\mathbf{x}=(x_1,x_2,\ldots,x_n)\}$ with *edge ordinal* n-D points $\mathbf{s}=(s_1,s_2,\ldots,s_n)$ $\mathbf{e}=(e_1,e_2,\ldots,e_n)$ such that

$$\forall i \in I_o \; s_i \leq x_i \leq e_i \quad (7)$$

For instance, if $X_i$ has values,1,2,3,4,5 and $s_i$=2, and $e_i$=4 then n-D points with values 2,3, or 4 of $X_i$ will be in this hyper-block

Definition. A **nominal hyper-block** is a set of nominal n-D points $\{\mathbf{x}=(x_1,x_2,\ldots,x_n)\}$ such that

$$\forall i \in I_n \; x_i \in Q_i \quad (8)$$

where $Q_i$ is a subset of values of attribute $X_i$. For instance, $X_i$ ={doctor, teacher, engineer} and $Q_i$ ={doctor, teacher}.

Definition. A **heterogeneous hyper-block** is a set of n-D points $\{\mathbf{x}=(x_1,x_2,\ldots,x_n)\}$ such that

$\forall i \in I_u \; \|x_i-c_i\| \leq L_i/2$ for all numeric attributes
$\forall i \in I_o \; s_i \leq x_i \leq e_i$ for all ordinal attributes and
$\forall i \in I_n \; x_i \in Q_i$ for all nominal attributes.

All numeric HBs are **interpretable** for data with numeric attributes of different modality types like temperature and blood pressure. HBs are interpretable because: (i) the distance in (4) is defined within each numeric attribute, where it is meaningful and (ii) do not include arithmetic operations between heterogenous attributes which are not meaningful, but only combine them with the meaningful logic operation. Thus, each such HB is a logical model, e.g., a HB contains all cases with temperature in [35,37] interval and blood pressure within [100,120] interval. Similarly, the hyperblocks for ordinal and nominal data are interpretable. As a result, this hyperblock for heterogeneous data is also interpretable.

Next larger interpretable heterogeneous datasets can be formed from smaller HBs. These small HBs serve as **interpretable atoms** for such larger datasets. We can formulate a stronger **conjecture** that **all** interpretable heterogeneous datasets with numeric, ordinal, and nominal attributes can be formed from HBs.

A hyperblocks is called a **pure HB** if it contains only cases of a single class. For every n-D point **x** it is possible to find a single class HB if there is no n-D point **y=x** that belongs for another class. Such HB can contain only n-D point **x**. Another advantage of numeric HBs is that they can be **visualized losslessly**, which is shown in [3].

The practice of coding the values of a nominal attribute consecutive integers 1,2,…,n is *contradictory* from stating that (1) it can **always** be used to encode nominal attributes without any limitation to stating that (2) this coding should **never** be used because it is not interpretable.

In fact, it should not be used in ML algorithms which conduct arithmetic operations with these codes like subtraction, squaring and so on. It is common in ML algorithms that use distances between n-D points like kNN. The use of this coding with the ML algorithms that do not make arithmetic operations with values of nominal attributes does not create interpretability problems. The logical algorithms which we outlined above belong to this category.

Another advantage of creating and visualizing logical rules is that often they are used as an efficient *tool to explain* deep learning and other black box models by mimicking behavior of these models.

### A. Proposed Toolkit

The toolkit includes the **Data Type Editor** integrated with visualization system **VisCanvas** [2,3] for multidimensional data visualization based on the adjustable parallel coordinates. The data type editor supports saving data in the interpretable measurement coding format for pattern discovery and data visualization. Fig. 1 illustrates setting up and applying a coding scheme for data that converts letter grades for 4 classes $X_1$-$X_4$ as follows: A to 4, B to 3, C to 2, and so on.

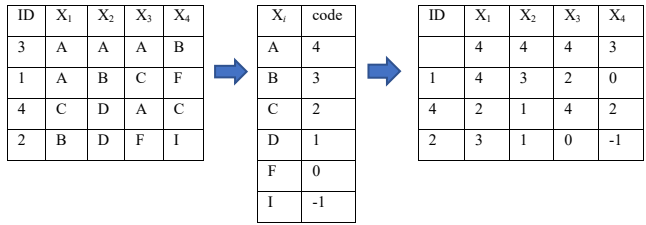

    original data        coding scheme        coded data

Fig. 1. Example of applying coding schema for class grades.

The toolkit supports Nominal, Ordinal, Interval, and Ratio data measurement types. The attributes of the absolute measurement data type are encoded as the ratio data type because it differs from it only by presence of the fixed measurement unit. The data type editor provides helpful descriptions and examples in case the user is not familiar with these data types. The user interface allows a user: to assign data type for each attribute, and to group values of each attribute. The **scheme loader** allows a user to assign data measurement type: nominal, ordinal, interval, and ratio.

A typical example of **heterogeneous data** are the mushroom data [9], which contain 8124 instances and 22 attributes. These attributes include **nominal** data such as habitat (grass, leaves, meadows, paths, urban, waste, woods), **ordinal** data, such as gill size (broad, narrow) and gill spacing (crowded, close, distant), **absolute** data such as the number of rings (0,1,2), which the scheme loader treats as **ratio** data. The colors of different parts of the mushroom such as cap color represent an interesting data type. It can be treated as: (1) nominal (red, blue, green and so on), (2) three numeric attributes like R,G,B, (3) some scalar function from R,G,B like R+G+B, and (4) a single numeric **ratio** data type that uses wavelength.

The first one does not capture the similarity between colors, the second one is expanding the number of attributes, the third one corrupts similarity relations between some colors, and the last one is the most physically meaningful. Since, each color covers a wavelength interval, grouping wavelength values according to colors is a natural way to encode the colors. These groups are ordered and can be encoded by integers starting from 0. In general, grouping attribute values decreases the space size and run time of the algorithms.

Manual coding is time consuming and tedious work for the tasks with many attributes and multiple values of each attribute. The toolkit allows to speed up this process. The editor has the "All Ordinal" and "All Nominal" options that allow to assign initially Nominal or Ordinal type and to assign integer code values from 1 to *n* to all attributes with abilities to edit this assignment later.

**Grouping**. Fig. 2 illustrates grouping and binary coding keys for a **nominal** attribute. Fig. 3 illustrates setting up groups for the numeric interval and ratio attributes by creating **intervals** where user sets on the starting value for the group and the length of its interval. Original values of the attributes may not correctly represent its data type for the task. The user can select keeping the existing values or generate new ones.

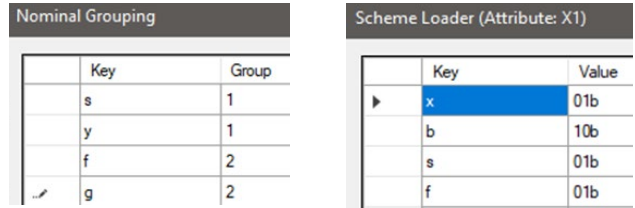

Fig. 2. Grouping keys and assigning binary codes for a nominal attribute.

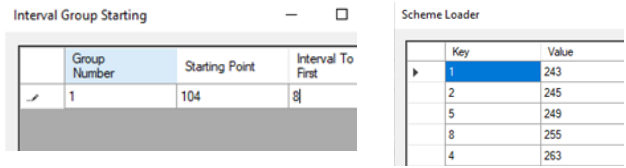

(a) Setting up intervals scheme.    (b) Resulting generated scheme.

Fig.3. Setting up groups for the numeric interval and ratio attributes.

**Hierarchy of attribute groups**. When we have hundreds of attributes, a hierarchy of attributes allows to deal with them efficiently. The system supports a user in constructing a hierarchy and picking up a level at which attributes will be visualized.

## IV. MIXED MULTIDIMENSIONAL DATA VISUALIZATION

This section presents methods to visualize heterogeneous multidimensional data in parallel coordinates implemented in the Toolkit.

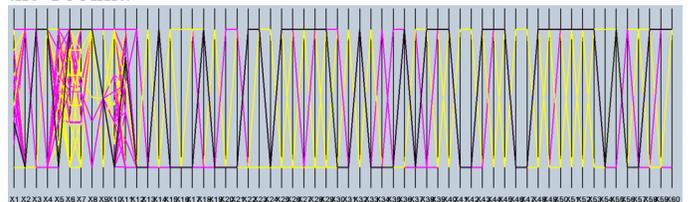

Fig.4. Encoded mushroom data in parallel coordinates.

Fig. 4 shows the mushroom data visualized in parallel coordinates in VisCanvas [2,3], where nominal attributes are encoded by binary attributes using the Data Type Editor described above resulted in over 70 attributes. The uninformative attributes with the same values for all cases are omitted leaving 60 attributes shown in Fig. 4. These 60 attributes from 22 original attributes demonstrate deficiency of such coding for visualization. The data are quite hard to look at since there are so many attributes. Moreover, in general, binary attributes do not fit well to be visualized in parallel coordinates because binary attributes do not have much variability and do not fill the area between 0 and 1 in each coordinate. As a result, many lines cover each other making different classes practically undistinguishable as was shown before in [10].

Fig. 5 shows visualization of the Census income dataset from the UCI ML repository [9] in parallel coordinates. It was selected for its size of 48842 cases of mixture of 14 integer and categorical attributes. Conversion several nominal attributes to binary attributes produced total 26 attributes. Binary coding of nominal attributes requires domain knowledge to make it meaningful, which is easier when attributes are a part of the common knowledge like in the Census data.

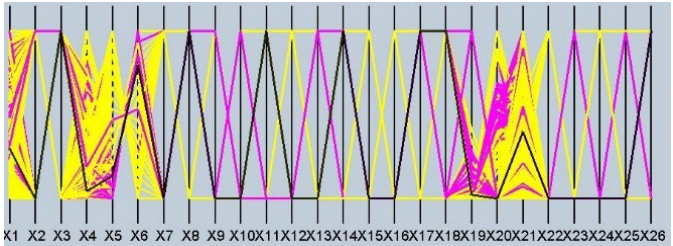

Fig. 5. Visualization in parallel coordinates of the Census income data with mixed data after applying meaningful coding.

Figs. 4, 5 and 6 show the deficiency of direct visualization of data with multiple nominal attributes by converting to many binary attributes. Fig. 6 visualizes the Teaching Assistant (TA) evaluation dataset from the UCI ML repository [9], which is a mix of categorical and integer data. In these data values of some attributes are grouped and converted to binary attributes using the Data Type Editor. The cases are colored according to the values of the attribute X9, which has over 20 values. It shows high overlap of cases making visual discovering of patterns very difficult. Therefore, below we present alternative methods.

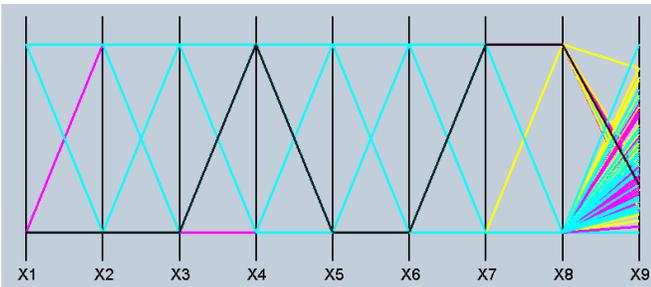

Fig. 6. Parallel coordinates visualization of TA data.

**Frequency based visualization of nominal attributes**. Fig. 7 shows the visualization of the same TA dataset in parallel coordinates where bars are sized by the **frequency** of occurrence of the respective nominal values. The user interface in VisCanvas allows a user to generate this frequency-based visualization. This visualization is similar to the **parallel sets** visualization [12] with thinner lines making visualization less occluded. The frequencies of values of X9 also have been computed and visualized by the height of the bars. The bars are ordered according to their frequencies in the descending order with bars of the greatest frequency at the bottom of X9. This visualization is more informative than shown in Fig. 6, which helps to solve the occlusion problem that we see in Fig. 6. Also, a user can select a specific color scheme for the nominal blocks.

**Reference frequency visualization**. The frequency visualizations shown in Fig. 7 adapted from parallel sets [12] have an important deficiency for machine learning tasks. They are *not related to the classes,* the heigh of the bar for a given value of the attribute is based on the *frequency* of this value *itself* in the dataset, which contain cases of several classes.

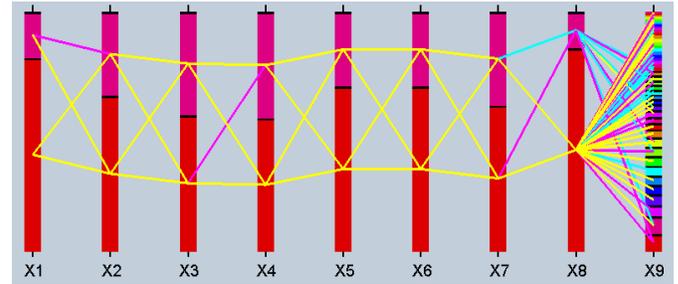

Fig. 7. Parallel sets inspired visualization of TA data.

Below we present visualizations that are more relevant to machine learning. This method computes the number of bars, and their heights based on the **relations** of these values with values of another **reference attribute** $X_t$ (e.g., target attribute). The coloring of lines and bars is based on the values of the reference attribute/classes. For instance, reference attribute can have 00 cases with $x_i$=a can contain 10 cases with $x_t$=0, 70 cases with $x_t$=1, 12 cases with $x_t$=2, and 8 cases with $x_t$=3. Here $x_t$=1 is a dominant value of $X_i$.

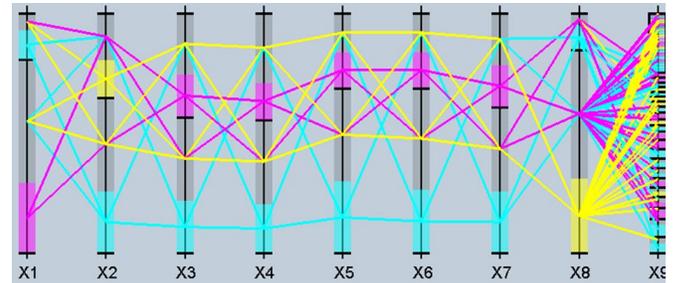

Fig. 8. Visualization with class colors and dominant frequency for TA dataset with weighted lines.

The bars for all non-dominant values can be joined to a single bar (see grey bars in Fig. 8). It visually emphasizes the dominant value of the target attribute/class. In Fig. 8 the portion of each bar is colored by the dominant class (magenta or blue) and the non-dominant part is grey, the black horizontal lines separate bars. In addition, wider lines allow a user to see the larger frequency of cases between bars. In attribute X9 (class size/number of students) all smaller frequencies under a threshold are put in one block at the top. In comparison with Fig. 6, it is now easier to see how many lines are going to each bar and understand dominance of classes.

Figs. 9 and 10 illustrate advantages of reference frequency-based visualization on the mushroom classification dataset with class colors. Fig. 9 shows dominant class frequency with all bars, while Fig. 10 shows only bars of high purity (≥80%).

The frequency approach implemented in these visualizations differs from frequency methods listed in [4]. The **Frequency Encoding** in [4] converts textual data into numeric data by assigning the frequency of that value as its code. In this coding if two colors get the same frequency, say, 0.3, then it will be used as code for both colors making them indistinguishable. In our frequency visualization these two colors will have their own bars of length 0.3 each. So, they will not collapse to a single bar and the information will be preserved. To distinguish our frequency encoding from described in [4] we will call our frequency encoding as a **visual frequency encoding**. The **Mean Encoding**/Target Encoding [4] has the same issue. It can produce the equal codes and glue values, while our visualization avoids the information loss.

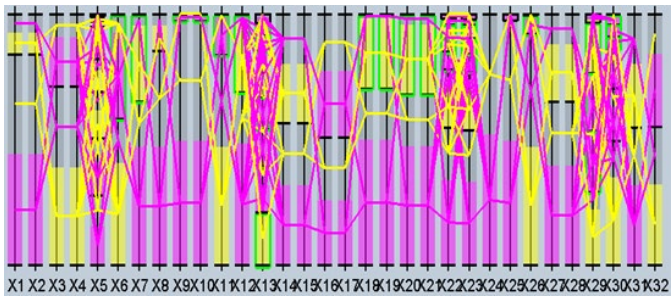

Fig. 9. Mushroom dataset with class colors and dominant frequency with all bars.

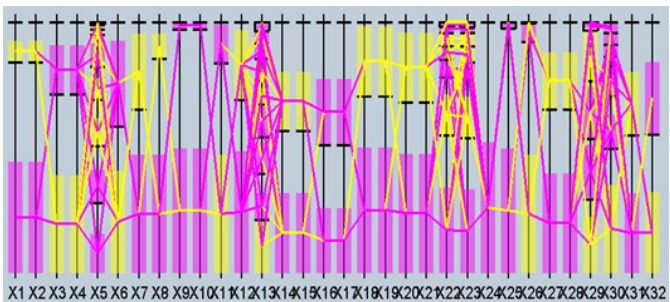

Fig. 10. Mushroom dataset with class colors and dominant frequency with all bars and with bars above 80% purity threshold.

To help a user to get more information from the visualization their linguistic descriptions are automatically generated (Fig. 11). It describes blocks/bars with purity of 80% or higher and attributes with small frequency blocks/bars. Fig. 11 describes a few attributes of the mushroom dataset visualized in Fig. 10. On this dataset many attributes have small blocks with purity above 80% purity. To emphasize larger blocks the limit on block size of 10% was introduced in addition to 80% of purity.

In Fig. 8, attribute X9 was not very useful in determining rules. In contrast, Fig. 12 shows the TA dataset with splitting values of attribute X9 (class size) into 4 groups. Now it shows the pattern that TAs rated the best (blue blocks and lines) were dominant in the first and the fourth groups. The worst rated TAs were dominant in the second group and the average TAs were dominant in the third group. **Flipping attributes** allows making visual patterns clearer and VisCanvas supports it. Fig. 13 shows flipping (negating) some attributes in the TA dataset. All attributes are normalized to [0,1] and flipping creates 1-x for the attribute x. As a result, blue cases concentrate at the bottom, magenta cases are in the middle and yeallow cases split between the top and the middle.

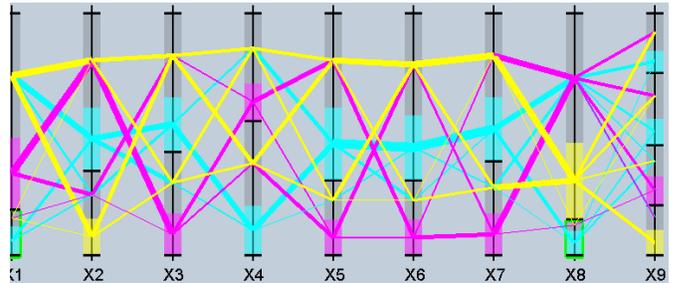

Fig.11. Dominant blocks/bars on Mushroom dataset.

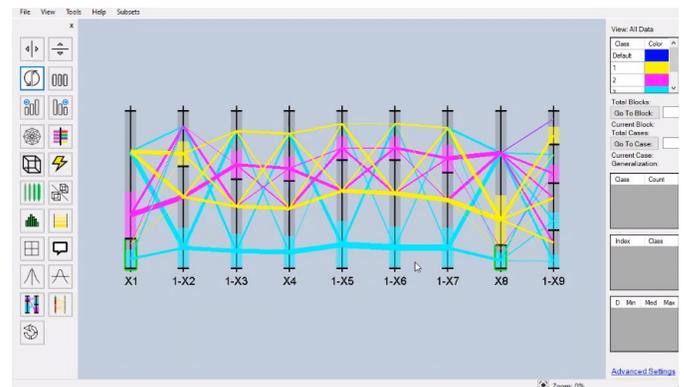

Fig. 12. TA dataset with blocks sorted by purity and wider frequent cases.

Fig. 13. TA dataset with flipped attributes to simplify visuals.

**Reordering attributes** is another option in VisCanvas to make patters clearer. Fig. 14 shows the mushroom data with reordered attributes, where now the attributes with the most pure blocks are on the left and the least pure blocks are on the right. It makes visual patters clearer. The user can change the purity threshold interactively for more distilled vusal patterns.

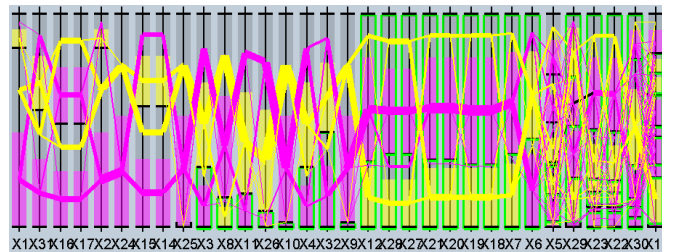

Fig. 14. Mushroom data with attributes ordered by purity of blocks decreasing from left to right and wider frequent cases.

**Grouping, relocating blocks**, **and sorting by color** is another way to make patterns simpler and clearer. Some datasets have attributes with a few large blocks and many small blocks. Often these small blocks are of high purity and are next to the large blocks, which making them hard to see as in Fig. 14 on the right. Therefore, the *smaller blocks* (under 20%) are moved to the top of those attributes (see Fig. 15). Fig. 15 also shows the results of *sorting* the mushroom data by *color* and putting the yellow blocks on the top. Here the attributes are sorted by the number of purity blocks.

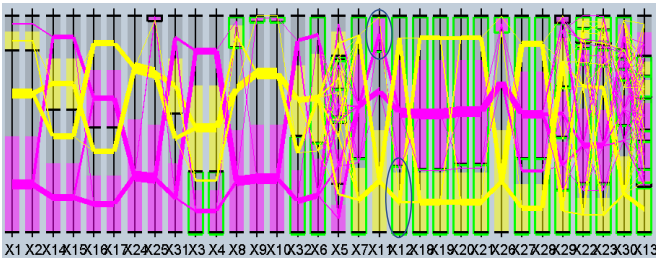

Fig, 15. Mushroom dataset sorted by color.

**Visual rule generation**. Such strong visual patterns in the data allow to make classification rules like

If case **a** is in magenta block in X11 & is NOT
in yellow block X12 then **a** is in yellow class     (7)

These blocks are outlined by the black ovals in Fig. 15. These magenta and yellow boxes have green frames indicating their high purity. This rule is an example of interpretable rules that end users can find themselves by such visual knowledge discovery process. The comparison of Fig. 15 with Fig. 4 of the same data shows the benefit of the toolkit. It is very clear that extracting rules like (7) is impossible in Fig. 4.

**The benefits of visual knowledge discovery** derived from visualization in Fig. 13 are described below for the TA dataset. This dataset is a collection of attributes of students who were rated 1 to 3 on their effectiveness of being a teaching assistant. The value 3 is a 'great' rating, 2 is an 'average' rating, and 1 is a 'bad' rating. In Fig. 13, 3 is blue, 2 is yellow, and 1 is magenta. After flipping X2-X7, and X9 we can see a clear pattern in the bottom of Fig. 13 that leads to a teaching assistant being great.

The attribute X1 shows a purity of over 60% dominant class 3 (great) in the bottom block and a dominant class of class 1(bad) in the top block. This attribute represents a Boolean value of whether the teaching assistant was a native English speaker or not where the top block is false, and the bottom block is true. This means that a teaching assistant is more likely to be rated as 'great' if they are native English speakers.

The first block in X2 is dominantly blue with most "great" TAs going there. It represents the value 0 meaning that a majority of student's label "great" had instructors from either group 1, 2, or 3. The upper block in the ovals with magenta dominant block is much less pure. Similarly, in X3 most of the "great" teaching assistants either had instructor in group 1 or group 4. Next, value 1 of the blue block in X4 means that the "great" teaching assistants came from group 1 or 3.

The analysis of attributes X5, X6, and X7 shows that most "bad" rated TAs were assisting in classes from group 4 since the first block value is one. The attribute X8 represents whether the semester was a normal semester or a summer semester. The bottom block of this coordinate is dominantly blue. Its value 2 means that most "great" teaching assistants were helping in classes during summer semesters. Fig. 13 shows that the bottom two blocks are dominantly blue in attribute X9 (class size). This means that "great" teaching assistants dominantly helped in classes with 3 to 19 students or 37 to 66 students.

## V. CONCLUSION AND FUTURE WORK

This paper presented the classification and the analysis of types of heterogeneous/mixed data in enhance interpretable Machine Learning. We proposed the adaptation of these mixed data types and their coding schemas for lossless visualization of multidimensional mixed data in parallel coordinates. The developed experimental Toolkit combines the Data Types Editor and VisCanvas system for mixed multidimensional data visualization and interpretable rule discovery. It is available on GitHub: https://github.com/SamShissler/VisCanvas2.0. It supports (1) numeric *coding schemes* for non-numeric attributes for *accurate and interpretable* ML with mixed data, (2) *lossless visualization* of n-D *non-numeric* data, and (3) *visual rule discovery* in these visualizations. The future work is developing new *full scope ML algorithms* for mixed data integrated with lossless visualization of n-D heterogeneous data to other types of General Line Coordinates.